# Improving Energy Efficiency in Manufacturing: A Novel Expert System Shell

Borys Ioshchikhes[*,a], Michael Frank[a], Tresa Maria Joseph[a], Matthias Weigold[a]

[a]*Technical University of Darmstadt, Institute for Production Management, Technology and Machine Tools (PTW), Otto-Berndt-Str. 2, 64287 Darmstadt, Germany*

[*] Corresponding author. *E-mail address*: b.ioschhikhes@ptw.tu-darmstadt.de

**Abstract**

Expert systems are effective tools for automatically identifying energy efficiency potentials in manufacturing, thereby contributing significantly to global climate targets. These systems analyze energy data, pinpoint inefficiencies, and recommend optimizations to reduce energy consumption. Beyond systematic approaches for developing expert systems, there is a pressing need for simple and rapid software implementation solutions. Expert system shells, which facilitate the swift development and deployment of expert systems, are crucial tools in this process. They provide a template that simplifies the creation and integration of expert systems into existing manufacturing processes. This paper provides a comprehensive comparison of existing expert system shells regarding their suitability for improving energy efficiency, highlighting significant gaps and limitations. To address these deficiencies, we introduce a novel expert system shell, implemented in Jupyter Notebook, that provides a flexible and easily integrable solution for expert system development.

*Keywords:* sustainability; energy analysis; knowledge-based system; climate neutrality; artificial intelligence

## 1. Introduction

As part of the European Green Deal, the European Union (EU) has committed to reducing greenhouse gas emissions by at least 55 % by 2030, compared to 1990 levels [1]. The industrial sector plays a pivotal role in this context, being responsible for approximately 30 % of greenhouse gas emissions, with energy use accounting for 24.2 %, making it the primary source of emissions [2]. Among various decarbonization efforts, increasing the share of electricity across all sectors is a key strategy. Projections indicate a rise in electricity's share of final energy consumption from 20 % currently to over 50 % by 2050. In particular, the industrial sector was the largest consumer of electric energy in 2019, comprising 41.9 % of the total consumption [3].

In response to these challenges, the revised Energy Efficiency Directive (EED) of the EU significantly raises the EU's ambition to enhance energy efficiency across various sectors, including industry. The EED gives legal standing to the fundamental EU energy policy principle of 'energy efficiency first', requiring EU countries to consider energy efficiency in all relevant policies and major investments [4]. However, despite regulatory efforts, several barriers hinder the improvement of energy efficiency in industry. These barriers include a shortage of specialized contractors, insufficient digital skills in the workforce for data analysis, and uncertainty about how to improve energy efficiency [5]. Expert systems (ESs) emerge as a potential solution to overcome these barriers. These systems can automatically identify energy efficiency potentials in manufacturing by analyzing energy data, pinpointing inefficiencies, and recommending optimizations that reduce energy consumption [6]. Although systematic approaches for developing such ESs exist [7, 8], there is an urgent need for streamlined software tools, known as expert system shells (ESSs), that provide a template that simplifies the creation of ESs [9]. Thus facilitating simple and rapid implementation.

Following the introduction, section 2 elaborates the concept of ESs and describes the role that ESSs play in their development. Section 3 provides a comprehensive evaluation of existing ESSs regarding their suitability for improving energy efficiency in manufacturing. This review identifies significant gaps and limitations in their applicability, highlighting areas where current solutions fall short. To address these deficiencies, section 4 presents a novel ESS. Finally, section 5 concludes with a summary and an outlook.

## 2. Background

*2.1. Expert system fundamentals*

ESs are computer programs designed to assist in decision-making and problem-solving in a particular domain [11]. Unlike conventional applications, ESs emulate human reasoning by representing human knowledge and utilize heuristic methods [12]. These heuristics, structured as IF-THEN rules, represent cause-and-effect relationships, with the IF part indicating causes (antecedents) and the THEN part indicating effects (consequents) [13].

Within the scope of improving energy efficiency in manufacturing, ESs can leverage expert knowledge and analyze data to automatically identify potentials [6]. ESs are comprised of several key components, as outlined by [13, 10] and illustrated in Fig. 1:

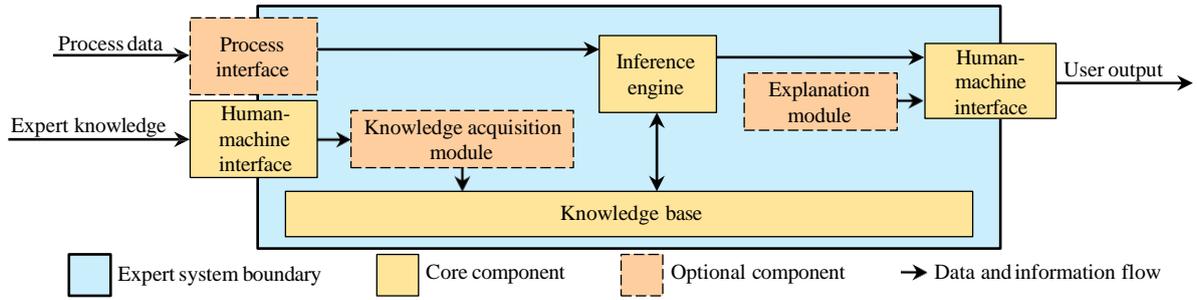

Fig. 1. Expert system architecture based on [10].

- **Knowledge base:** Stores expert knowledge and can be separated into short-term and long-term memory. The long-term memory holds heuristic rules from human experts, while the short-term memory functions like a database, storing or removing facts used by these rules.
- **Inference engine:** By leveraging the knowledge stored in the knowledge base, the inference engine emulates the reasoning of human experts. To solve complex problems or generate conclusions, it aligns facts from short-term memory with rules from long-term memory.
- **Human-machine interface:** Acts as the communication interface between the expert or user and the ES.
- **Process interface:** Enables the connection to other technical systems for unidirectional or bidirectional data transfer.
- **Explanation module:** Explains the reasoning carried out by the inference engine, making it understandable for the user and thereby enhancing its credibility and acceptance.
- **Knowledge acquisition module:** Allows the long-term memory of the knowledge base to be updated with new content, even after the ES has been deployed.

*2.2. Expert system development*

To develop ESs for improving energy efficiency in manufacturing, we present a systematic approach in [8] based on the Design Science Research Methodology by [14]. The development proceeds in several steps: Initially, energy-relevant machines and information are identified, followed by the definition of energy performance indicators (EnPIs). The EnPIs are then used to set a rule base that mirrors the decision-making process of human experts. A fuzzy rule base is suitable for this purpose because it can deal with partially true or false conditions, which reflects the way human experts make decisions without precise information [15]. Within the development process, measurement data is acquired to implement algorithms or create data-driven models that automate time series analysis for calculating EnPIs. As development progresses, artifacts are generated, representing either standalone components like algorithms, models, a fuzzy rule base, or the complete ES built from these elements. The artifacts are then applied and validated. If they reveal functional deficiencies or issues with inherent properties, further refinement iterations are required.

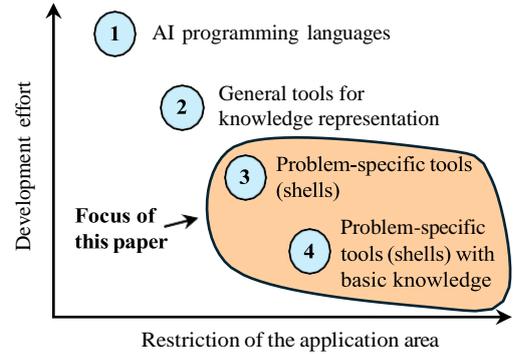

Fig. 2. Classification of tools for expert systems based on [9].

Four personas, each with distinct skills and expertise, support this development process. Machine operators contribute crucial practical experience, ensuring the feasibility of proposed optimizations. Energy managers assess manufacturing processes from an energy perspective, analyzing energy consumption and identifying inefficiencies. Knowledge engineers structure and integrate domain expertise into the ES, while data scientists design data-driven models and algorithms to automate the analysis performed by energy managers.

*2.3. Expert system tools*

[10] indicate that software implementations of ESs for improving energy efficiency are highly heterogeneous, relying on a diverse range of programs and programming languages. Appropriate ES tools can generalize and formalize established concepts, thereby accelerating and simplifying the development process. Regarding development effort and application area, these tools can be categorized into the following four groups (see Fig. 2) [9]:

1. Artificial intelligence (AI) programming languages, such as the logic-based language Prolog.
2. General tools for knowledge representation, such as rules with forward chaining.
3. Problem-specific tools (shells) that provide specific knowledge representations and problem-solving strategies.
4. Problem-specific tools (shells) that possess basic application knowledge and only require supplementation with domain-specific knowledge.

As shown in Fig. 2, the effort required to develop ESs decreases with a narrower application area of ES tools. For this reason, this paper focuses on the two latter groups. In general, ESSs can be defined as software containing a collection of functions that enable developers to build ESs. Selecting an ESS with the capabilities best suited to a particular problem can streamline system development and result in a more efficient solution [16].

## 3. Review of expert system shells

The findings of [10] suggest that no ESSs for improving energy efficiency in manufacturing exist or have not yet been successfully established. In this section, we therefore describe criteria and evaluate their degree of fulfillment for existing ESSs. These criteria are based on [16, 17, 18] and supplemented by current development trends as described in [10]. The evaluation of ESSs is similar to other software, but additionally includes unique criteria. Unlike procedural programming, where knowledge is implicitly embedded, ESs use declarative approaches with explicit knowledge stored in structures like rules. Therefore, evaluation criteria must account for these knowledge representation schemes [16]. The evaluated ESSs were selected from the implementations reviewed in [10] and the results of an exploratory search for tools that meet or exceed group 2 in Fig. 2. Table 1 additionally features our novel Expert System Shell for Energy Efficiency (ESS4EE), which is presented in more detail in section 4. The evaluation criteria are grouped into five categories based on [16].

The end-user interface is decisive for the usability and acceptance of software. For ESs, the following criteria are especially relevant:

- Result visualization [16, 17]: Built-in functions for data visualization displaying results, enabling clear graphical representation of information.
- User interactivity [17]: Users can engage with and control the system for re-execution or making changes.
- Documentation availability [16, 18]: Comprehensive guides and reference materials for users.
- User training and support [16, 18]: Access to tutorials, training programs, and technical assistance for users.

Together with the end user interface, the development interface is the second human-centered category. It is crucial for building the ES and is divided into the following criteria:

- Programming language popularity [16]: Widely used and highly demanded programming language according to [19].
- Integrability [16]: Ability to connect and interact with external programs, databases, and other systems.
- Extensibility [16, 18]: Ability to be expanded or modified to accommodate new features, knowledge representations, or functionalities.
- User-friendly debugging [18]: Provides intuitive and powerful debugging tools to detect problems or inconsistencies in rule sets that may result in runtime errors.
- Ease of learning [16, 18]: Measure of difficulty for a new user to understand and use the system.

The next two evaluation categories address technical aspects. The system interface is essential for integrating with other software and hardware components and involves the following criteria:

- Portability [16, 18]: Ability of the ESS to operate across different platforms and environments with minimal modifications.
- Embeddability [16, 18]: Refers to the ability to integrate the ESS into other software applications or systems.
- Scalability [18]: Capacity to handle increasing amounts of data or more complex rules without significant performance loss.
- Security [16]: Enables the system to be protected against unauthorized access, data breaches and vulnerabilities.

The second technical evaluation category focuses on core system capabilities. The ES functionality is fundamental for the classification outlined in Fig. 2 and encompasses the subsequent criteria:

- Knowledge base modularization [18]: Capability to organize the knowledge base into separate, manageable modules or components.
- Efficient rule handling [16]: Efficiency and manner in which the ES processes and applies rules to reach conditions.
- Forward chaining [16]: Provides the reasoning method, which starts with known facts and applies inference rules to derive new conclusions.
- Response to complex scenarios [16]: Capability to manage and resolve intricate multi-step scenarios.
- Machine learning (ML) model integration [10]: Offers interfaces for integrating ML models for automated data analysis.
- Process interface [17]: Provides interfaces for the connection of live machine and sensor data, e.g. via OPC UA or Modbus.
- Real-time data processing [16, 18]: Ability to continuously read, process, and analyze data as it is generated, providing immediate insights with minimal latency.
- Uncertainty handling [16, 17, 18]: Ability to deal with non-binary uncertainties in information and decision-making processes, e.g. using fuzzy logic.

The final evaluation category, cost-related criteria, addresses economic considerations:

- Simple installation [16]: Simplicity of installing and configuring the system to make it fully operational.
- Open source availability [10]: Characteristic of software where the source code is freely accessible to the public.
- Low licensing cost [16]: Systems that require minimal fees for acquisition and maintenance of legal usage rights, ensuring affordability for individuals or organizations.
- Community support [17]: Assistance, guidance, and resources offered by a collective of users, developers, or enthusiasts, without the need to purchase any books or materials.

The evaluation in Table 1 highlights a gap in the availability of ESSs tailored to improving energy efficiency in manufacturing. In particular, the ES functionalities are insufficiently fulfilled by existing tools.

## 4. Expert system shell for energy efficiency (ESS4EE)

In response to the shortcomings identified through the systematic evaluation of existing EESs in section 3, we develop the Expert System for Energy Efficiency (ESS4EE), which is available in a GitHub repository [24]. Its implementation is carried out within Jupyter Notebook, a web application for the creation and exchange of notebooks, code, and data [25]. The code used in our shell is written in Python. The ESS4EE is structured around a modular software architecture, as illustrated in Fig. 3, which allows for flexibility in adapting to various manufacturing environments, scalability to handle different industrial scenarios of varying complexity and integration with external systems.

In its default state, the ESS4EE can be assigned to group 3 according to the classification in Fig. 2, as it provides specific knowledge representations and problem-solving strategies. For example, the ESS4EE includes predefined components for EnPI definition, fuzzy inference system (for handling uncertainties with fuzzy logic), and ML model integration. These features enable the system to address energy data analysis and optimization tasks in manufacturing without requiring users to build the logic or representation structures from scratch. However, the ESS4EE can furthermore be associated with group 4 by adding predefined application knowledge, such as standardized EnPIs or industry-specific energy information.

The system architecture consists of several key components, each playing a distinct role in ensuring efficient knowledge management, data processing, and decision-making. The *KnowledgeBase* contains persistent expert knowledge provided by the knowledge engineer. This knowledge includes detailed descriptions of machinery, relevant data points, energy-related information, EnPIs, a rule base, and example datasets. Additionally, the *KnowledgeBase*

Table 1. Comparison of expert system shells.

| | CLIPS v6.4 [20] | Jess v8.1 [21] | Drools v8.43 [22] | PyKE v1.1.1 [23] | ESS4EE [24] |
|---|---|---|---|---|---|
| **End-user interface** | | | | | |
| Data and result visualization | ○ | ○ | ● | ○ | ● |
| User interactivity | ○ | ◐ | ● | ◐ | ● |
| Documentation availability | ● | ● | ● | ● | ◐ |
| User training and support | ◐ | ◐ | ● | ◐ | ◐ |
| **Developer interface** | | | | | |
| Programming language popularity | ◐ | ● | ● | ● | ● |
| Integrability | ◐ | ● | ● | ◐ | ● |
| Extensibility | ◐ | ● | ● | ◐ | ● |
| User-friendly debugging | ◐ | ● | ● | ◐ | ◐ |
| Ease of learning | ◐ | ◐ | ◐ | ● | ● |
| **System interface** | | | | | |
| Portability | ◐ | ● | ● | ● | ● |
| Embeddability | ● | ● | ● | ● | ● |
| Scalability | ◐ | ● | ● | ◐ | ● |
| Security | ○ | ◐ | ● | ◐ | ◐ |
| **ES functionality** | | | | | |
| Knowledge base modularization | ● | ● | ● | ◐ | ● |
| Efficient rule handling | ● | ● | ● | ◐ | ● |
| Forward chaining properties | ● | ● | ● | ● | ● |
| Response to complex scenarios | ● | ● | ● | ◐ | ● |
| ML model integration | ○ | ○ | ◐ | ○ | ● |
| Process interface | ○ | ○ | ◐ | ○ | ● |
| Real-time data processing | ○ | ○ | ◐ | ○ | ◐ |
| Uncertainty handling | ◐ | ◐ | ◐ | ◐ | ● |
| **Cost-related criteria** | | | | | |
| Simple installation | ● | ● | ● | ● | ● |
| Open source availability | ● | ○ | ● | ● | ● |
| Low licensing cost | ● | ◐ | ● | ● | ● |
| Community support | ◐ | ◐ | ● | ◐ | ○ |

● Fully fulfilled  ◐ Partly fulfilled  ○ Not fulfilled can be modified and enriched with temporary knowledge related to specific use cases, which can be input by the user through the *UserInterface*. Measurement and process data from the production environment can be accessed via the *Connectors* provided by the *ProcessInterface*. This enables not only historical data but also live data from power meters and programmable logic controllers to be analyzed. Support functions are provided within the *Helpers* component to facilitate the integration of algorithms and ML models, providing advanced analytics capabilities. This component also handles the assignment of addresses to data points and includes visualization functions. The *InferenceEngine* is responsible for applying the fuzzy inference system. It features membership functions for input and output variables, an *InferenceMechanism*, and a machine-readable *FuzzyRuleBase*. Together, these elements enable the system to process data and information to provide actionable insights. The inputs to this system are the machine-readable EnPIs, which are defined in the *KnowledgeBase*. For their computation, both predefined calculation methods and analytics using algorithms and ML models can be applied. The results from the *InferenceEngine* are presented to the user with visualizations, which, together with intermediate outputs, contribute to the *Explanation* component, improving the system's interpretability.

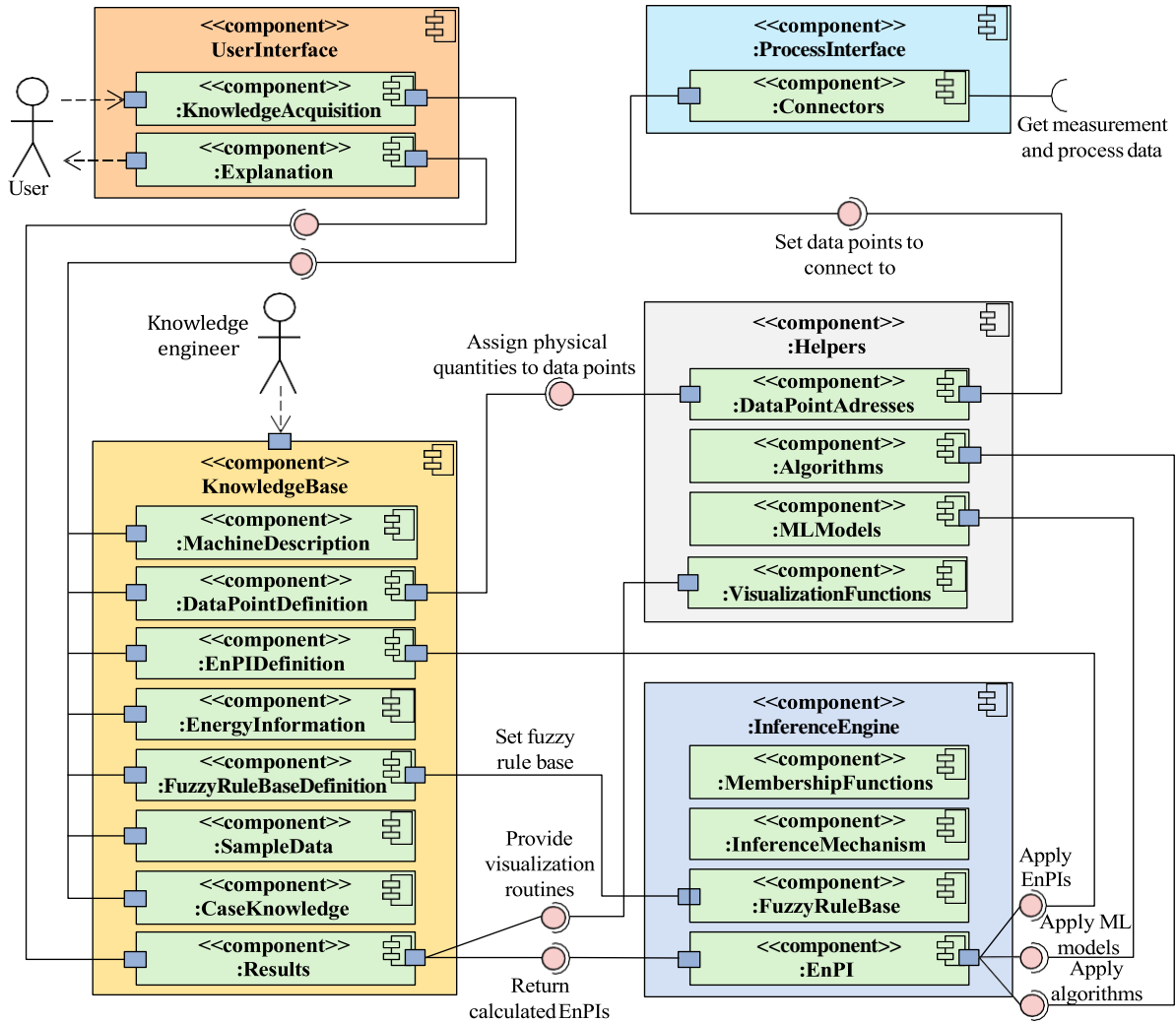

Fig. 3. Component diagram of the ESS4EE.

## 5. Summary and outlook

This paper presents a comprehensive evaluation of tools for developing ESs with a focus on improving energy efficiency in manufacturing. Through this evaluation, we identified several shortcomings in existing solutions, particularly concerning ES functionalities. In response, we developed our novel ESS4EE. Compared to currently available solutions, the ESS4EE offers tailored functionalities specifically designed to optimize energy efficiency in manufacturing environments. Key features include its process interface, which enables data access to the production environment, real-time data processing, and the ability to manage uncertainties through fuzzy logic. Moreover, the ESS4EE is the only tool to offer a structured framework that supports the systematic development of ESs within this domain. It includes problem-specific components such as templates for machine descriptions and the definition of EnPIs, which significantly streamline and accelerate the software development process. However, despite its improvements, the ESS4EE lacks the comprehensive documentation, user training, and support that established tools offer. Additionally, there is no community of users outside the development team to provide assistance or contribute to the tool's growth. Addressing these limitations will be critical to expanding the adoption of the ESS4EE in broader industrial context.

The outlook for the ESS4EE includes several key steps for further development. In the near term, we plan to develop multiple ESs across diverse industrial use cases to demonstrate the ESS4EE's adaptability and effectiveness. These implementations will be accompanied by user studies to assess the system's usability and identify areas for improvement. Furthermore, we aim to explore the integration of large language models (LLMs) to assist with knowledge acquisition and formalization within the ES. By leveraging LLMs, we can reduce the dependency on knowledge engineers, automating the process of capturing and structuring expert knowledge, thereby enhancing the system's scalability and ease of use. In the long term, fostering a supportive user community will be essential for the continued evolution of the ESS4EE. A robust community would facilitate the sharing of best practices, contribute to documentation efforts, and provide user support, helping to address the current gaps in training and assistance. These steps will contribute to the ESS4EE having a significant impact on energy efficiency in manufacturing and serve as a vital tool in this area.

**CRediT author statement**

**Borys Ioshchikhes**: Conceptualization, Methodology, Investigation, Software, Writing - Original Draft, Visualization. **Tresa Maria Joseph**: Conceptualization, Software, Writing - Original Draft, Formal analysis. **Michael Frank**: Conceptualization, Methodology, Resources, Writing - Original Draft. **Matthias Weigold**: Supervision, Project administration, Funding acquisition.

**References**

[1] European Parliament, Council of the European Union, Regulation (eu) 2021/1119 (30.06.2021).
URL https://eur-lex.europa.eu/legal-content/EN/TXT/?uri=CELEX%3A32021R1119

[2] H. Ritchie, M. Roser, P. Rosado, $CO_2$ and greenhouse gas emissions (2020).
URL https://ourworldindata.org/co2-and-greenhouse-gas-emissions

[3] International Energy Agency, Key world energy statistics 2021 (2021).
URL https://www.iea.org/reports/key-world-energy-statistics-2021/final-consumption

[4] European Parliament, Council of the European Union, Directive (eu) 2023/1791 (13.09.2023).
URL https://eur-lex.europa.eu/legal-content/EN/TXT/?uri=OJ%3AJOL_2023_231_R_0001&qid=1695186598766

[5] Accelerating ambition: How global industry is speeding up investment in energy efficiency (2022).
URL https://www.energyefficiencymovement.com/wp-content/uploads/2022/04/ABB-Energy-Efficiency-Survey-Report-2022.pdf

[6] B. Ioshchikhes, G. Elserafi, M. Weigold, An expert system-based approach for improving energy efficiency of chamber cleaning machines, in: Proceedings of the Conference on Production Systems and Logistics (CPSL 2023), publish-Ing., Offenburg, 2023, pp. 1–11. doi:10.15488/13419.

[7] B. Ioshchikhes, M. Weigold, Development of stationary expert systems for improving energy efficiency in manufacturing, Procedia CIRP 126 (2024) 921–926. doi:10.1016/j.procir.2024.08.351.

[8] B. Ioshchikhes, M. Frank, G. Elserafi, J. Magin, M. Weigold, Developing expert systems for improving energy efficiency in manufacturing: A case study on parts cleaning, Energies 17 (14) (2024) 3417. doi:10.3390/en17143417.

[9] F. Puppe, Einführung in Expertensysteme, 2nd Edition, Studienreihe Informatik, Springer, Berlin and Heidelberg, 1991.

[10] B. Ioshchikhes, M. Frank, M. Weigold, A systematic review of expert systems for improving energy efficiency in the manufacturing industry, Energies 17 (19) (2024) 4780. doi:10.3390/en17194780.

[11] A. W. DeTore, An introduction to expert systems, Journal of insurance Medicine 21 (4) (1989) 233–236.

[12] P. Jackson, Introduction to Expert Systems, 3rd Edition, Addison-Wesley Longman Publishing Co., Inc, USA, 1998.

[13] G. P. Buccieri, J. A. P. Balestieri, J. A. Matelli, Energy efficiency in brazilian industrial plants: knowledge management and applications through an expert system, Journal of the Brazilian Society of Mechanical Sciences and Engineering 42 (11) (2020). doi:10.1007/s40430-020-02667-x.

[14] Hevner, March, Park, Ram, Design science in information systems research, MIS Quarterly 28 (1) (2004) 75. doi:10.2307/25148625.

[15] S.-H. Liao, Expert system methodologies and applications—a decade review from 1995 to 2004, Expert Systems with Applications 28 (1) (2005) 93–103. doi:10.1016/j.eswa.2004.08.003.

[16] A. C. Stylianou, G. R. Madey, R. D. Smith, Selection criteria for expert system shells, Communications of the ACM 35 (10) (1992) 30–48. doi:10.1145/135239.135240.

[17] A. Martin, R. K. Law, Expert system for selecting expert system shells, Information and Software Technology 30 (10) (1988) 579–586. doi:10.1016/0950-5849(88)90114-0.

[18] J. Rothenberg, Jody Paul, Iris Kameny, James R. Kipps, Marcy Swenson, Evaluating Expert System Tools: A Framework and Methodology, RAND Corporation, Santa Monica, CA, 1987.

[19] L. S. Vailshery, Most widely utilized programming languages among developers worldwide 2024 (2024).
URL https://www.statista.com/statistics/793628/worldwide-developer-survey-most-used-languages/

[20] Clips: A tool for building expert systems (2023).
URL https://www.clipsrules.net/

[21] E. J. Friedman-Hill, Jess, the java expert system shell (1997). doi:10.2172/565603.
URL https://www.osti.gov/biblio/565603,journal=

[22] Drools (2023).
URL https://www.drools.org/

[23] B. Frederiksen, Applying expert system technology to code reuse with pyke (2008).
URL https://pyke.sourceforge.net

[24] B. Ioshchikhes, T. M. Joseph, Expert System Shell for Energy Efficiency (ES4EE) (Apr. 2025).
URL https://github.com/Borika95/ES4EE

[25] F. Loizides, B. Schmidt, Thomas Kluyver, Benjamin Ragan-Kelley, Fernando Pérez, Brian Granger, Matthias Bussonnier, Jonathan Frederic, Kyle Kelley, Jessica Hamrick, Jason Grout, Sylvain Corlay, Paul Ivanov, Damián Avila, Safia Abdalla, Carol Willing (Eds.), Jupyter Notebooks – a publishing format for reproducible computational workflows, 2016.